\documentclass[letterpaper, 10pt]{article}

\usepackage[utf8]{inputenc}
\usepackage[T1]{fontenc}

\usepackage{spconf}
\usepackage[american]{babel}

\usepackage{amsmath}
\usepackage{amsfonts}
\usepackage{mathtools}

\usepackage{epsfig}
\usepackage{hyperref}

\usepackage{ragged2e}
\usepackage{ifthen}
\usepackage[left=1.8cm, right=1.8cm, top=1.5cm, bottom=4cm]{geometry}
\usepackage{fancyhdr}
\usepackage{tikz}
\usepackage{tkz-berge}
\usetikzlibrary{fit, shapes, arrows.meta, automata, backgrounds, calendar} 
\definecolor{colorPrimary}              {RGB}{0, 0, 0}
\definecolor{colorSecondary}            {RGB}{100,100,100}
\definecolor{colorDeactivated}          {RGB}{100,100,100}
\definecolor{colorTertiary}             {RGB}{0,75,90}
\colorlet{colorTertiary33}{colorTertiary!33!white}
\definecolor{colorQuaternary}           {RGB}{203, 81, 25}
\colorlet{colorQuaternary33}{colorQuaternary!33!white}
\definecolor{colorQuinary}              {RGB}{ 36,143,133}
\definecolor{colorSenary}               {RGB}{  0,145,168}
\definecolor{colorBackgroundPrimary}    {RGB}{255,255,255}
\definecolor{colorBackgroundDeactivated}{RGB}{255,255,255}
\definecolor{colorBackgroundSecondary}  {RGB}{255,255,255}
\definecolor{colorBackground}           {RGB}{255,255,255}

\definecolor{colori}  {rgb}{0.0 , 0.0 , 0.0 }
\definecolor{colorii} {rgb}{0.16, 0.31, 0.44}
\definecolor{coloriii}{rgb}{0.94, 0.59, 0.36}
\definecolor{coloriv} {rgb}{0.41, 0.54, 0.30}
\definecolor{colorv}  {rgb}{0.41, 0.16, 0.30}
\definecolor{halfgray}{rgb}{0.4 , 0.4 , 0.4 }
\definecolor{gray20}  {rgb}{0.2 , 0.2 , 0.2 }
\definecolor{gray40}  {rgb}{0.4 , 0.4 , 0.4 }
\definecolor{gray60}  {rgb}{0.6 , 0.6 , 0.6 }
\definecolor{gray80}  {rgb}{0.8 , 0.8 , 0.8 }

\newcommand{\colorDeactivated}[1]  {{\color{colorDeactivated}  {#1}}}

\usepackage{tikz}
\usepackage{tkz-berge}
\usepackage{smartdiagram}
\usepackage{pgfplots}
\usepackage{pgfplotstable}
\usepgfplotslibrary{fillbetween}

\usetikzlibrary{positioning,fit,shapes,arrows.meta,automata,backgrounds,calendar,datavisualization,datavisualization.formats.functions,fadings} 
\usetikzlibrary{spy}

\tikzfading[name=fade right,
  left color=transparent!0  , right color=transparent!100]
\tikzfading[name=fade left,
  left color=transparent!100, right color=transparent!0]
\tikzfading[name=fade right25,
  left color=transparent!0  , right color=transparent!75]
\tikzfading[name=fade left25,
  left color=transparent!75, right color=transparent!0]
\tikzfading[name=fade right33,
  left color=transparent!0  , right color=transparent!66]
\tikzfading[name=fade left33,
  left color=transparent!66, right color=transparent!0]
\tikzfading[name=fade right50,
  left color=transparent!0  , right color=transparent!50]
\tikzfading[name=fade left50,
  left color=transparent!50, right color=transparent!0]

\tikzset{
  System/.style = { rectangle,
	                  draw = colorPrimary,
										fill = colorBackgroundPrimary,
										inner sep = 2pt,
										outer sep = 0pt,
										minimum size = 20pt,
										font = \normalsize,
										scale = 0.75
									},
  Signal/.style = { -{latex},
								  thin,
									color = colorPrimary
								},
}

\tikzset{
  FactorNode/.style = { rectangle,
	                      draw = colorPrimary,
												fill = colorBackgroundPrimary,
												inner sep = 2pt,
												outer sep = 0pt,
												minimum size = 20pt,
												font = \normalsize,
												scale = 0.75
												},
  OperationNode/.style = { rectangle,
	                      draw = colorPrimary,
												fill = colorBackgroundPrimary,
												inner sep = 2pt,
												outer sep = 0pt,
												minimum size = 15pt,
												font = \normalsize,
												scale = 0.75
												},
  VariableNode/.style = { ellipse,
	                      draw = colorPrimary,
												fill = colorBackgroundPrimary,
												inner sep = 2pt,
												outer sep = 0pt,
												minimum size = 15pt,
												font = \normalsize,
												scale = 0.75
												},
  PriorNode/.style = { rectangle,
	                     draw = colorPrimary,
										   fill = colorBackgroundPrimary,
											 inner sep = 2pt,
											 outer sep = 0pt,
											 minimum size = 15pt,
											 font = \scriptsize,
											 scale = 0.75
										 },
  EmptyNode/.style = { rectangle,
											 inner sep = 2pt,
											 outer sep = 0pt,
											 minimum size = 20pt,
											 font = \normalsize,
											 scale = 0.75
										 },
  DotNode/.style = { rectangle,
											 inner sep = 0pt,
											 outer sep = 0pt,
											 minimum size = 0pt,
											 font = \normalsize,
											 scale = 0.75
										 },
  NodeLabelAbove/.style = { above = 0.375em,
			    			            font  = \scriptsize
					                },
  NodeLabelBelow/.style = { below = 0.375em,
			    			            font  = \scriptsize
					                },
  NodeLabelLeft/.style  = { left = 0.375em,
			    			            font = \scriptsize
					                },
  NodeLabelRight/.style = { right = 0.375em,
			    			            font  = \scriptsize
					                },
  NodeLabelAboveRight/.style = { above right = 0.375em,
			    			                 font  = \scriptsize
					                     },
  NodeLabelAboveLeft/.style = { above left = 0.375em,
			    			                font  = \scriptsize
					                    },
  NodeLabelBelowRight/.style = { below right = 0.375em,
			    			                 font  = \scriptsize
					                     },
  NodeLabelBelowLeft/.style  = { below left = 0.375em,
			    			                 font = \scriptsize
					                     },
  Edge/.style = { -{latex},
								  thin,
									color = colorPrimary
								},
  MsgArrow/.style = { -{latex},
											thin,
											color = colorSecondary
								    },
  EdgeNoArrow/.style = { thin,
									       color = colorPrimary
								       },
  HighlightEdge/.style = { -{Latex[length=2mm,width=4mm]},
								           line width=1mm,
													 draw = colorSecondary
                         },
  HighlightEdgeSmall/.style = { -{Latex[length=1mm,width=2mm]},
								           line width=0.44mm,
													 draw = colorSecondary
                         },
  HighlightEdgeBig/.style = { -{Latex[length=5mm,width=8mm]},
								           line width=5mm,
													 draw = colorSecondary
                         },
  Label/.style = { midway,
									 font = \scriptsize
								 },
	HighlightBlock/.style = { fill = colorSecondary,
														draw = colorSecondary,
														opacity = 0.25,
														dash dot,
	                          inner sep = -1pt,
													  rectangle,
													  draw
													},
  Axes/.style = { -{latex},
								  thin,
									color = colorPrimary
								},
  PlotAnnotation/.style = { -{latex},
	                          thin,
									          color = colorQuaternary
								},
  PlotAnnotationNoArrow/.style = { thin,
									          color = colorQuaternary
								},
  PlotHighlight/.style = { -{latex},
	                          thin,
									          color = colorSecondary
								},
  PlotHighlightNoArrow/.style = { thin,
									                color = colorSecondary
								},
  AxesTicks/.style = { font  = \scriptsize,
											 color = colorPrimary,
					           },
}

\tikzset{
  FactorNodeDeactivated/.style = { rectangle,
	                      draw = colorDeactivated,
												text = colorDeactivated,
												opacity = 0.5,
												fill = colorBackgroundDeactivated,
												inner sep = 2pt,
												outer sep = 0pt,
												minimum size = 20pt,
												font = \normalsize,
												scale = 0.75
												},
  OperationNodeDeactivated/.style = { rectangle,
	                      draw = colorDeactivated,
												text = colorDeactivated,
												opacity = 0.5,
												fill = colorBackgroundDeactivated,
												inner sep = 2pt,
												outer sep = 0pt,
												minimum size = 15pt,
												font = \normalsize,
												scale = 0.75
												},
  PriorNodeDeactivated/.style = { rectangle,
	                     draw = colorDeactivated,
											 opacity = 0.5,
										   fill = colorBackgroundDeactivated,
											 inner sep = 2pt,
											 outer sep = 0pt,
											 minimum size = 15pt,
											 font = \scriptsize,
											 scale = 0.75
										 },
  EmptyNodeDeactivated/.style = { rectangle,
											 opacity = 0.5,
											 inner sep = 2pt,
											 outer sep = 0pt,
											 minimum size = 20pt,
											 font = \normalsize,
											 scale = 0.75
										 },
  NodeLabelAboveDeactivated/.style = { above = 0.375em,
												    opacity = 0.5,
			    			            font  = \scriptsize
					                },
  NodeLabelBelowDeactivated/.style = { below = 0.375em,
												    opacity = 0.5,
			    			            font  = \scriptsize
					                },
  NodeLabelLeftDeactivated/.style  = { left = 0.375em,
												    opacity = 0.5,
			    			            font = \scriptsize
					                },
  NodeLabelRightDeactivated/.style = { right = 0.375em,
												    opacity = 0.5,
			    			            font  = \scriptsize
					                },
  EdgeDeactivated/.style = { -latex,
								  thin,
								  opacity = 0.5,
									color = colorDeactivated
								},
  HighlightEdgeDeactivated/.style = { -{Latex[length=2mm,width=4mm]},
												   draw opacity = 0.5,
								           line width=1mm,
                         },
  LabelDeactivated/.style = { midway,
									 font = \scriptsize,
									 opacity = 0.5,
									 color = colorDeactivated
								},
	HighlightBlockDeactivated/.style = { fill = black!20,
														dash dot,
	                          inner sep = 2pt,
													  rectangle,
													  draw opacity = 0.5
													},
}

\tikzset{
  SmallFactorNode/.style = { rectangle,
	                      draw = colorPrimary,
												fill = colorBackgroundPrimary,
												inner sep = 2pt,
												outer sep = 0pt,
												minimum size = 12pt,
												font = \normalsize,
												scale = 0.75
												},
  SmallOperationNode/.style = { rectangle,
	                      draw = colorPrimary,
												fill = colorBackgroundPrimary,
												inner sep = 2pt,
												outer sep = 0pt,
												minimum size = 7.5pt,
												font = \scriptsize,
												scale = 0.75
												},
  SmallVariableNode/.style = { ellipse,
	                      draw = colorPrimary,
												fill = colorBackgroundPrimary,
												inner sep = 2pt,
												outer sep = 0pt,
												minimum size = 7.5pt,
												font = \normalsize,
												scale = 0.75
												},
  SmallPriorNode/.style = { rectangle,
	                     draw = colorPrimary,
										   fill = colorBackgroundPrimary,
											 inner sep = 2pt,
											 outer sep = 0pt,
											 minimum size = 7.5pt,
											 font = \scriptsize,
											 scale = 0.75
										 },
  SmallEmptyNode/.style = { rectangle,
											 inner sep = 2pt,
											 outer sep = 0pt,
											 minimum size = 10pt,
											 font = \normalsize,
											 scale = 0.75
										 },
  SmallDotNode/.style = { rectangle,
											 inner sep = 0pt,
											 outer sep = 0pt,
											 minimum size = 0pt,
											 font = \normalsize,
											 scale = 0.75
										 },
  SmallNodeLabelAbove/.style = { above = 0.375em,
			    			            font  = \scriptsize
					                },
  SmallNodeLabelBelow/.style = { below = 0.375em,
			    			            font  = \scriptsize
					                },
  SmallNodeLabelLeft/.style  = { left = 0.375em,
			    			            font = \scriptsize
					                },
  SmallNodeLabelRight/.style = { right = 0.375em,
			    			            font  = \scriptsize
					                },
  SmallEdge/.style = { -{latex},
								  thin,
									color = colorPrimary
								},
  SmallEdgeNoArrow/.style = { thin,
									       color = colorPrimary
								},
  SmallHighlightEdge/.style = { -{Latex[length=2mm,width=4mm]},
								           line width=1mm,
													 draw = colorSecondary
                         },
  SmallLabel/.style = { midway,
									 font = \scriptsize
								},
	SmallHighlightBlock/.style = { fill = black!20,
														dash dot,
	                          inner sep = 2pt,
													  rectangle,
													  draw
													},
  SmallAxes/.style = { -{latex},
								  thin,
									color = colorPrimary
								},
  SmallAxesTicks/.style = { font  = \scriptsize,
											 color = colorPrimary,
					                },
}

\tikzset{
  SmallFactorNodeDeactivated/.style = { rectangle,
	                      draw = colorQuaternary,
												text = colorQuaternary,
												opacity = 0.5,
												fill = colorBackgroundSecondary,
												inner sep = 2pt,
												outer sep = 0pt,
												minimum size = 12pt,
												font = \normalsize,
												scale = 0.75
												},
  SmallOperationNodeDeactivated/.style = { rectangle,
	                      draw = colorQuaternary,
												text = colorQuaternary,
												opacity = 0.5,
												fill = colorBackgroundSecondary,
												inner sep = 2pt,
												outer sep = 0pt,
												minimum size = 7.5pt,
												font = \scriptsize,
												scale = 0.75
												},
  SmallPriorNodeDeactivated/.style = { rectangle,
	                     draw = colorQuaternary,
											 opacity = 0.5,
										   fill = colorBackgroundSecondary,
											 inner sep = 2pt,
											 outer sep = 0pt,
											 minimum size = 7.5pt,
											 font = \scriptsize,
											 scale = 0.75
										 },
  SmallEmptyNodeDeactivated/.style = { rectangle,
											 opacity = 0.5,
											 inner sep = 2pt,
											 outer sep = 0pt,
											 minimum size = 10pt,
											 font = \normalsize,
											 scale = 0.75
										 },
  SmallNodeLabelAboveDeactivated/.style = { above = 0.375em,
												    opacity = 0.5,
			    			            font  = \scriptsize
					                },
  SmallNodeLabelBelowDeactivated/.style = { below = 0.375em,
												    opacity = 0.5,
			    			            font  = \scriptsize
					                },
  SmallNodeLabelLeftDeactivated/.style  = { left = 0.375em,
												    opacity = 0.5,
			    			            font = \scriptsize
					                },
  SmallNodeLabelRightDeactivated/.style = { right = 0.375em,
												    opacity = 0.5,
			    			            font  = \scriptsize
					                },
  SmallEdgeDeactivated/.style = { -latex,
								  thin,
								  opacity = 0.5,
									color = colorQuaternary
								},
  SmallHighlightEdgeDeactivated/.style = { -{Latex[length=2mm,width=4mm]},
												   draw opacity = 0.5,
								           line width=1mm,
                         },
  SmallLabelDeactivated/.style = { midway,
									 font = \scriptsize,
									 opacity = 0.5,
									 color = colorQuaternary
								},
	SmallHighlightBlockDeactivated/.style = { fill = black!20,
														dash dot,
	                          inner sep = 2pt,
													  rectangle,
													  draw opacity = 0.5
													},
}

\tikzset{
  PrimaryDataPlot/.style = { color = colorPrimary,
												     line width=0.75pt,
											     },
  SecondaryDataPlot/.style = { color = colorSecondary,
												     line width=0.75pt,
											     },
  TertiaryDataPlot/.style = { color = colorTertiary,
												     line width=0.75pt,
											     },
  QuaternaryDataPlot/.style = { color = colorQuaternary,
												     line width=0.75pt,
											     },
  QuinaryDataPlot/.style = { color = colorQuinary,
												     line width=0.75pt,
											     },
  SenaryDataPlot/.style = { color = colorSenary,
												     line width=0.75pt,
											     },
  PrimaryDataPlotTransparent/.style = { color = colorPrimary,
																					opacity = 0.5,
												     line width=0.75pt,
											     },
  SecondaryDataPlotTransparent/.style = { color = colorSecondary,
																					opacity = 0.5,
												     line width=0.75pt,
											     },
  TertiaryDataPlotTransparent/.style = { color = colorTertiary,
																					opacity = 0.5,
												     line width=0.75pt,
											     },
  QuaternaryDataPlotTransparent/.style = { color = colorQuaternary,
																					opacity = 0.5,
												     line width=0.75pt,
											     },
  QuinaryDataPlotTransparent/.style = { color = colorQuinary,
																					opacity = 0.5,
												     line width=0.75pt,
											     },
  SenaryDataPlotTransparent/.style = { color = colorSenary,
																					opacity = 0.5,
												     line width=0.75pt,
											     },
}

\pgfplotsset{compat=1.8}
\pgfplotsset{PrimaryAxes/.style= { line width=0.5pt,
																	 axis background style = {fill = colorBackgroundPrimary},
																	 major grid style = {dotted, colorDeactivated, line width=0.5pt},
																	 grid = major,
																   axis lines = left, 
																	 scaled ticks = false,
																	 color = colorPrimary,
																	 axis x line = box, 
																	 axis y line = box,
																	 axis line/.style = {-latex},
																 	 font = \scriptsize,
																	 legend style = {%
																			shape = rectangle, 
																			draw = colorDeactivated, 
																			fill = colorBackgroundPrimary,
																			line width = 0.5pt,
																			font = \scriptsize,
																			cells = {anchor = west}, 
																			inner xsep = 3pt,
																			inner ysep = 2pt,
																			nodes = {inner sep = 2pt, text depth = 0.15em}, 
																	 },
								                 }}

	\pgfplotsset{select coords between index/.style 2 args={
			x filter/.code={
					\ifnum\coordindex<#1\fi
					\ifnum\coordindex>#2\fi
			}
	}}

\usepackage{mleftright}
\usepackage{cleveref}

\newcommand*{\R}{\mathbb{R}}

\newcommand{\func}[2]{\mathop{{}#1} \mleft( #2 \mright)}
\newcommand{\sfunc}[2]{\mathop{{}#1} ( #2 )}
\newcommand{\prob}[1]{\sfunc{p}{ #1 }}
\newcommand*{\normalN}{\mathcal{N}}
\newcommand{\normaldist}[1]{\func{\normalN}{#1}}
\newcommand{\snormaldist}[1]{\sfunc{\normalN}{#1}}
\newcommand*{\di}[0]{\, \mathrm{d}}
\newcommand{\Int}[4]{\int\limits_{#1}^{#2} #4 \di #3}
\newcommand{\Expectation}[1]{\mathop{{}\mathrm{E}} \mleft[ #1 \mright]}
\newcommand{\Covariance}[1]{\mathop{{}\mathrm{Cov}} \mleft[ #1 \mright]}
\newcommand{\identity}{\mathcal{I}}

\newcommand{\fwd}[1]{\overrightarrow{#1}}
\newcommand{\bwd}[1]{\overleftarrow{#1}}
\newcommand*{\px}{\prob{x}}
\newcommand*{\py}{\prob{y}}
\newcommand*{\pyx}{\prob{y|x}}
\newcommand*{\fx}{\sfunc{f}{x}}
\newcommand*{\gx}{\sfunc{g}{x}}
\newcommand*{\mean}{m}

\newcommand*{\dimx}{n}
\newcommand*{\dimy}{m}
\newcommand*{\spacex}{\R^\dimx}
\newcommand*{\spacey}{\R^\dimy}
\newcommand*{\npoints}{\ell}
\newcommand*{\sigmapoint}{\xi}
\newcommand*{\weight}{w}
\newcommand*{\degreef}{k}

\newcommand*{\covmat}{C}
\newcommand*{\varmat}{V}

\newcommand*{\infomat}{W}

\newcommand*{\processnoise}{w}
\newcommand*{\measnoise}{v}
\newcommand*{\Processnoise}{W}
\newcommand*{\Measnoise}{V}

\title{On Approximate Nonlinear Gaussian Message Passing on Factor Graphs}
\name{Eike Petersen, Christian Hoffmann, and Philipp Rostalski}
\address{\small Institute for Electrical Engineering in Medicine, University of Lübeck, Germany. \\ 
\small E-mail adresses: \{eike.petersen, christian.hoffmann, philipp.rostalski\}@uni-luebeck.de \\
}
\begin{document}
%
\maketitle
\begin{abstract}
Factor graphs have recently gained increasing attention as a unified framework for representing and constructing algorithms for signal processing, estimation, and control.
One capability that does not seem to be well explored within the factor graph tool kit is the ability to handle deterministic nonlinear transformations, such as those occuring in nonlinear filtering and smoothing problems, using tabulated message passing rules.
In this contribution, we provide general forward (filtering) and backward (smoothing) approximate Gaussian message passing rules for deterministic nonlinear transformation nodes in arbitrary factor graphs fulfilling a Markov property, based on numerical quadrature procedures for the forward pass and a Rauch-Tung-Striebel-type approximation of the backward pass.
These message passing rules can be employed for deriving many algorithms for solving nonlinear problems using factor graphs, as is illustrated by the proposition of a nonlinear modified Bryson-Frazier (MBF) smoother based on the presented message passing rules.
\end{abstract}

\section{Introduction}
\label{sec:intro}
Factor graphs are a graphical representation of statistical independence statements regarding probability distributions.
These probabilistic graphical models facilitate the application of inference algorithms by means of message passing on the graph~\cite{Kschischang2001a, Kschischang2001, Loeliger2004, Loeliger2007}. 
Many classical algorithms such as recursive least squares (RLS)~\cite{Loeliger2007}, linear Kalman filtering and smoothing~\cite{Wadehn2016}, and expectation maximization~\cite{Dauwels2009} have already been formulated as message passing on a factor graph.
This enables the simple generation of novel algorithms by adapting the factor graph of known algorithms to a given problem, or by combining multiple known algorithms into a single graph and, hence, a joint algorithm.
Algorithms adapted to a variety of problems such as cooperative localization~\cite{Li2015}, sparse input estimation~\cite{Loeliger2016}, and motion planning~\cite{Dong2016} have been derived.

Despite the vast amount of literature on nonlinear Gaussian filtering and smoothing algorithms~\cite{Wan2000, Arasaratnam2007, Saerkkae2008, Arasaratnam2009, Jia2012}, to the authors' knowledge very few efforts have been made to represent existing nonlinear filtering algorithms in the factor graph framework.
Meyer, Hlinka, and Hlawatsch~\cite{Meyer2014} describe sigma-point belief propagation (SPBP) algorithms on factor graphs, but they consider a non-sequential system model with observations depending on pairs of states, hence making their results inapplicable to sequential filtering and smoothing problems.
Deisenroth and Mohamed~\cite{Deisenroth2012} propose the use of expectation propagation (EP) as a general framework for Gaussian smoothers, similarly to \cite{ZoHe05}, in which moments of distributions from the univariate exponential family are approximated using Gaussian quadrature.
Both approaches result in iterative update rules with respect to the marginals.

In the present paper, we make an attempt at providing concise update rules for performing approximate Gaussian message passing through deterministic nonlinear nodes in factor graphs, thus enabling the representation of known nonlinear filtering and smoothing algorithms, and facilitating the derivation of new algorithms for various nonlinear problems.
Similarly to~\cite{ZoHe05, Deisenroth2012}, we propose the use of numerical quadrature for computing moments, however we focus our exposition on deriving efficient approximate message passing rules for the directed messages instead of the marginals, which results in non-iterative filtering and smoothing schemes for graphs without loops.

\section{Nonlinear Transformations}
\label{sec:filtering}
Consider a deterministic, nonlinear transformation 
\begin{equation}
 	y = \fx \in \spacey \quad \Leftrightarrow \quad \prob{y|x} = \func{\delta}{y-\fx}
 	\label{eq:nonlin-trafo}
\end{equation}
of a random variable $X$ with values in $\spacex$.
The probability density function (PDF) of $Y$ is given by
\begin{equation}
	\py = \Int{\spacex}{}{x}{\pyx \px}
	= \Int{\spacex}{}{x}{\sfunc{\delta}{y-\fx}\px},
	\label{eq:pdf}
\end{equation}
the expected value of $Y$ results as
\begin{equation}
	\mean_Y \coloneqq \Expectation{Y} = \Int{\spacey}{}{y}{y \py}
	= \Int{\spacex}{}{x}{\fx \px}
	\label{eq:transformed-expectation}
\end{equation}
and the covariance as
\begin{multline}
	\varmat_Y = \infomat_Y^{-1} \coloneqq \Covariance{Y} = \Int{\spacey}{}{y}{(y-\mean_Y)(y-\mean_Y)^T\py} \\
	= \Int{\spacex}{}{x}{(\fx-\mean_Y)(\fx-\mean_Y)^T\px}.
	\label{eq:transformed-covariance}
\end{multline}
For arbitrary nonlinear functions $f$ and PDFs $\px$, the integrals in eqs. \eqref{eq:transformed-expectation} and \eqref{eq:transformed-covariance} rarely admit closed-form analytical solutions.
Hence, the prior density $\px$ is usually assumed to be normally distributed, and numerical quadrature procedures are employed to approximate both integrals, such as the unscented transform (UT)~\cite{Wan2000}, Gauss-Hermite quadrature (GHQ)~\cite{Arasaratnam2007}, spherical-radial transform (SRT), which in a filtering setting results in the Cubature Kalman Filter (CKF)~\cite{Arasaratnam2009}, or sparse-grid quadrature using the Smolyak rule~\cite{Jia2012}.
The resulting estimates for $\Expectation{Y}$ and $\Covariance{Y}$ are then used as the parameters of a normal approximation to the PDF $\py$, which amounts to approximate moment matching and, hence, approximate minimization of the Kullback-Leiber divergence between the actual distribution $\py$ and its normal approximation~\cite{Barber2012}.

All of the methods mentioned above have in common that they perform the approximation
\begin{equation}
	\Int{\spacex}{}{x}{\gx \normaldist{x; 0, \identity}} \approx \sum_{i=1}^{\npoints} \weight_i \sfunc{g}{\sigmapoint_i},
	\label{eq:numerical-integration}
\end{equation}
of a normally weighted integral, where the number $\npoints$ of integration points $\sigmapoint_i$ and the weights $\weight_i$ differ between methods, and $\normaldist{x; \mean, \varmat}$ denotes the PDF of a normally distributed variable $x$ with mean $\mean$ and covariance matrix $\varmat$.
Note that in order to solve eq.~\eqref{eq:transformed-covariance} using the approximation~\eqref{eq:numerical-integration}, one chooses
\begin{equation*}
	\gx \equiv (\fx - \mean_Y)(\fx-\mean_Y)^T,
\end{equation*}
and hence $\gx$ is a polynomial of degree $2\degreef$ if $f$ is a polynomial of degree $\degreef$.
As a consequence, both the UT and the SRT quadrature formulas, which yield exact results for polynomials up to and including degree $3$, do not calculate the covariance exactly for polynomials $f$ of order $2$~\cite{Arasaratnam2009}.
In particular, this includes bilinear functions $f$ resulting from a multiplication of two Gaussian random variables.
Gauss-Hermite quadrature formulas of arbitrary order can be readily constructed, but unfortunately these formulas suffer heavily from the curse of dimensionality~\cite{Arasaratnam2009}.
Sparse-grid quadrature rules on the other hand, of which the classical UT has been shown to be a subset, can be flexibly adjusted in their degree of precision  with a number of quadrature points growing polynomially in the number of dimensions, hence alleviating the curse of dimensionality~\cite{Jia2012}.

Having described a feasible strategy for performing the forwards (filtering) pass through a nonlinear transformation, we will now consider the backwards (smoothing) pass.
If the inverse~$f^{-1}$ of the nonlinear transformation is available, the same approach that has been described so far can also be used to implement the backward pass.
If, however, this is not the case, nonlinear Rauch-Tung-Striebel-type (RTS) smoothers can be derived, such as the unscented RTS smoother proposed by Särkkä~\cite{Saerkkae2008}.
In the following, a general nonlinear Gaussian RTS smoother-type backward pass through a nonlinear function node in a factor graph is derived, following the derivation in~\cite{Saerkkae2008} but employing a slightly more general setting to facilitate local interpretation in a factor graph context.

Consider again the deterministic nonlinear transformation~\eqref{eq:nonlin-trafo}, and assume the parameters $\fwd{\mean}_X, \fwd{\mean}_Y, \mean_Y, \fwd{\varmat}_X, \fwd{\varmat}_Y$, and $\varmat_Y$ of the filtering distributions
\begin{align}
	\prob{x|u} &\sim \normaldist{x; \fwd{\mean}_X,\fwd{\varmat}_X} \quad \text{and} \label{eq:def-filt}\\
	\prob{y|u} &\sim \normaldist{y; \fwd{\mean}_Y,\fwd{\varmat}_Y}
\end{align}
and the smoothing distribution of $y$
\begin{equation}
	\prob{y|u, z}\sim\snormaldist{y; \mean_Y,\varmat_Y}
\end{equation}
known, where $u$ denotes knowledge on $x$, and~$z$ denotes knowledge on~$y$.
The aim is now to compute the parameters $\mean_X, \varmat_X$ of a Gaussian approximation
\begin{equation}
	\prob{x|u,z}\sim\snormaldist{x; \mean_X,\varmat_X},
	\label{eq:def-smooth}
\end{equation}
to the smoothing distribution of $x$ incorporating all available data.
To this end, further assume that the model satisfies the Markov conditions
\begin{equation}
	\prob{z|x,y} = \prob{z|y} \quad \text{and} \quad \prob{y|u,x} = \prob{y|x}.
	\label{eq:markov}
\end{equation}
It then holds that 
\begin{gather*}
	\prob{x, y|u} = 
	\prob{x|u} \prob{y|x} \quad \text{and}\\
	\prob{x|u, y} = \prob{x|u, y, z}
\end{gather*}
due to the assumed Markov property~\eqref{eq:markov}.
It follows that
\begin{align*}
	\prob{x, y|u, z}
	&= \prob{x | u, y, z}
		\prob{y|u, z} \\
	&= \frac{\prob{x, y|u} \prob{y|u, z}}{
	\prob{y|u}}
\end{align*}
and by marginalization of $y$
\begin{equation}
	\prob{x|u, z} =
	\Int{\spacey}{}{y}{
		\frac{\prob{x, y|u} \prob{y|u, z}}{
	\prob{y|u}}
	}.
	\label{eq:conditional-marginalization}
\end{equation}
If the joint distribution $\prob{x, y|u}$ is approximated by a normal distribution
\begin{equation*}
	\prob{x, y|u} \sim \normaldist{\begin{pmatrix} x \\ y\end{pmatrix};
	\begin{pmatrix}
		\fwd{\mean}_{X} \\ \fwd{\mean}_{Y}
	\end{pmatrix},
	\begin{pmatrix}
		\fwd{\varmat}_{X} & \fwd{\covmat} \\
		\fwd{\covmat}^T & \fwd{\varmat}_{Y}
	\end{pmatrix}
	},
\end{equation*}
with suitably chosen covariance matrix $\fwd{\covmat}$, the marginalization~\eqref{eq:conditional-marginalization} can be evaluated analytically to~\cite{Saerkkae2008}
\begin{equation}
	\prob{x|u, z} \sim
	\normaldist{x; \mean_{X}, \varmat_{X}}
	\label{eq:rts-results}
\end{equation}
with $\mean_X$, $\varmat_X$, $\fwd{D}$, and $\fwd{\covmat}$ defined as in eqs.~\eqref{eq:rts-mean},~\eqref{eq:rts-cov},~\eqref{eq:D-def} and~\eqref{eq:covmat-approx} in \cref{tab:mp}, where $\fwd{\covmat}$ has been approximated using any of the quadrature methods described previously.

\section{Message passing on factor graphs}
\label{sec:factorgraphs}
\begin{figure}[tb]
	\centering
	\begin{tikzpicture}
		\def\xInc {1.125}
		\def\yInc {0.8}
		\SetGraphUnit{1}
		\node[EmptyNodeDeactivated]     (begin) at (-1.0, 0.0) {$\colorDeactivated{\ldots}$};
		\node[OperationNodeDeactivated] (e01)   at ($(begin)   + (\xInc, 0.0)$) {$\colorDeactivated{=}$};
		\node[FactorNodeDeactivated]    (C0)    at ($(e01)     - (0.0, 1.375*\yInc)$) {$\colorDeactivated{h}$};
		\node[OperationNodeDeactivated] (p01)   at ($(C0)      - (0.0, \yInc)$) {$\colorDeactivated{+}$};
		\node[PriorNodeDeactivated] (Nd0) at ($(p01)   - (\xInc, 0.0)$) {};
		\node at (Nd0) [NodeLabelAboveDeactivated] {$\colorDeactivated{\normalN_{\Measnoise_{i - 1}}}$}; 
		\node[PriorNodeDeactivated] (Nz0) at ($(p01)   - (0.0, \yInc)$) {}; 
		\node at (Nz0) [NodeLabelLeftDeactivated] {$\colorDeactivated{\normalN_{Y_{i - 1}}}$}; 

		\draw[EdgeDeactivated] (begin) -- (e01) node[LabelDeactivated, above] {};
		\draw[EdgeDeactivated] (e01)   -- (C0)  node[LabelDeactivated, right] {};
		\draw[EdgeDeactivated] (C0)    -- (p01) node[LabelDeactivated, right] {};
		\draw[EdgeDeactivated] (p01)   -- (Nz0) node[LabelDeactivated, right] {$\colorDeactivated{Y_{i  -  1}}$};
		\draw[EdgeDeactivated] (Nd0)   -- (p01) node[LabelDeactivated, below] {$\colorDeactivated{\Measnoise_{i  -  1}}$};
		
		\node[FactorNode]    (A1)     at ($(e01)   + (\xInc, 0.0)$)   {${f}$}; 
		\node[OperationNode] (p11)    at ($(A1)    + (\xInc, 0.0)$)   {${+}$};
		\node[OperationNode] (p12)    at ($(p11)   + (\xInc, 0.0)$)   {${+}$};
		\node[OperationNode] (e11)    at ($(p12)   + (\xInc, 0.0)$)   {${=}$};
		\node[FactorNode]    (B1)     at ($(p12)   + (0.0, \yInc)$)   {${g}$};
		\node[FactorNode]    (C1)     at ($(e11)   - (0.0, 1.375*\yInc)$)   {${h}$};
		\node[OperationNode] (p13)    at ($(C1)    - (0.0, \yInc)$)   {${+}$};
		\node[PriorNode] (Nd1) at ($(p13)   - (\xInc, 0.0)$) {};
		\node at (Nd1) [NodeLabelAbove] {$\normalN_{\Measnoise_{i}}$}; 
		\node[PriorNode] (Nu1) at ($(B1)   + (0.0, \yInc)$) {}; 
		\node at (Nu1) [NodeLabelRight] {$\normalN_{U_{i}}$}; 
		\node[PriorNode] (Nz1) at ($(p13)  - (0.0, \yInc)$) {}; 
		\node at (Nz1) [NodeLabelLeft] {$\normalN_{Y_{i}}$}; 
		\node[PriorNode] (Ni1) at ($(p11)  + (0.0, 2*\yInc)$) {}; 
		\node at (Ni1) [NodeLabelLeft] {$\normalN_{\Processnoise_{i}}$};

		\draw[Edge] (e01) -- (A1)  node[Label, above] {${X_{i -  1}}$};
		\draw[Edge] (A1)  -- (p11) node[Label, above] {};
		\draw[Edge] (p11) -- (p12) node[Label, above] {};
		\draw[Edge] (Nu1) -- (B1)  node[Label, right] {${U_{i}}$};
		\draw[Edge] (B1)  -- (p12) node[Label, right] {};
		\draw[Edge] (p12) -- (e11) node[Label, above] {};
		\draw[Edge] (e11) -- (C1)  node[Label, right] {};
		\draw[Edge] (C1)  -- (p13) node[Label, right] {};
		\draw[Edge] (p13) -- (Nz1) node[Label, right] {${Y_{i}}$};
		\draw[Edge] (Nd1) -- (p13) node[Label, below] {${\Measnoise_{i}}$};
		\draw[Edge] (Ni1) -- (p11) node[Label, left]  {${\Processnoise_{i  -  1}}$};
		
		\node[FactorNodeDeactivated]    (A2)    at ($(e11)   + (\xInc, 0.0)$) {$\colorDeactivated{f}$}; 
		\node[EmptyNodeDeactivated]     (end)   at ($(A2)    + (\xInc, 0.0)$) {$\colorDeactivated{\ldots}$};
 
		\draw[Edge] (e11) -- (A2) node[Label, above] {$X_{i}$};
		\draw[EdgeDeactivated] (A2) -- (end) node[LabelDeactivated, above] {};
		
\begin{scope}[on background layer]
	\node[inner sep=10pt, rectangle, dashed, draw = colorSecondary, fit=(A1) (e11) (Nz1) (Nu1)] {};
\end{scope}


\end{tikzpicture}

	\caption{One slice of a Forney-style factor graph representing the nonlinear state space model in \cref{eq:nonlinear-ss}. Capital letters indicate the random variables associated with edges.}
	\label{fig:kalman-fg}
\end{figure}
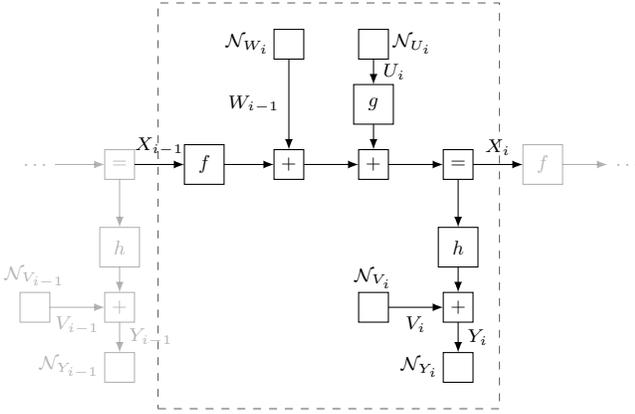

A factor graph is a graphical representation of a factorization of an arbitrary function.
Forney-style factor graphs (FFGs) consist of nodes, which represent factors, and edges connecting these nodes, which represent the variables that each factor depends on~\cite{Loeliger2007}.
Inference can be efficiently performed by means of message passing along the edges of the factor graph.
Edges are undirected, but arrows are introduced to disambiguate between messages $\overrightarrow{\mu}$ and $\overleftarrow{\mu}$ in and against the direction of an edge, respectively.
\Cref{fig:kalman-fg} shows an FFG representation of the PDF of the nonlinear state space model 
\begin{equation}
	\begin{aligned}
		x_i &= \func{f}{x_{i-1}} + \func{g}{u_{i}} + \processnoise_i \\
		y_i &= \func{h}{x_i} + \measnoise_i
	\end{aligned}
	\label{eq:nonlinear-ss}
\end{equation}
with deterministic nonlinear functions~$f$,~$g$, and~$h$, inputs~$u_i$ and process and measurement noise $\processnoise_i\sim \normaldist{0, \sigma_\Processnoise^2}$ and $\measnoise_i\sim\normaldist{0, \sigma_\Measnoise^2}$, respectively.
One of the main benefits of the factor graph framework is that it allows for the easy combination of existing algorithms from different fields, such as filtering/smoothing, sparse input estimation, parameter estimation, and control~\cite{Dauwels2009, Loeliger2016, Hoffmann2017, Hoffmann2017a} to derive powerful new algorithms.

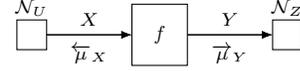
\begin{figure}[tb]
	\centering
	\begin{tikzpicture}[FactorNode/.append style = {inner sep = 11pt}]
		\def\xInc {1.7}
		\def\yInc {0.8}
		\SetGraphUnit{1}
		
		\node[PriorNode] (leftprior) at ($(0,0)   - (\xInc, 0.0)$) {};
		\node at (leftprior) [NodeLabelAbove] {$\normalN_U$}; 
		\node[FactorNode]    (f)     at ($(0,0)$)   {${f}$}; 
		\node[PriorNode] (rightprior) at ($(0,0)   + (\xInc, 0.0)$) {};
		\node at (rightprior) [NodeLabelAbove] {$\normalN_Z$};

		\draw[Edge] (leftprior) -- (f)  node[Label, above] {$X$};
		\draw[Edge] (f) -- (rightprior)  node[Label, above] {$Y$};
		\draw[Edge] (leftprior) -- (f)  node[Label, below] {$\overleftarrow{\mu}_X$};
		\draw[Edge] (f) -- (rightprior)  node[Label, below] {$\overrightarrow{\mu}_Y$};
	\end{tikzpicture}
	\caption{Factor node representing the factor $\func{\delta}{y-\fx}$.}
	\label{fig:nonlin-node}
\end{figure}

The dashed box in \cref{fig:kalman-fg} represents the joint probability $\prob{x_i, y_i, x_{i-1}, u_i}$, or equivalently 
the conditional distribution $\prob{x_i, y_i \lvert x_{i-1}, u_i} = \frac{\prob{x_{i}, y_{i}, x_{i-1}, u_{i}}}{\prob{x_{i-1}} \prob{u_i}}$.
Accordingly, for $N$ samples and initial state $x_{0}$ this results in the Markov chain
\begin{equation}
\begin{aligned}
&
p(y_1, \ldots, y_N, x_0, \ldots, x_N\,\vert\,u_1, \ldots, u_N)
\notag
\\
&\qquad =
	p(x_{0})\cdot\prod_{i=1}^{N}p(y_{i},x_{i}\,\vert\, x_{i-1},u_{i})
	\label{equ:ssm_jp}
\end{aligned}
\end{equation}
representing the PDF of the complete sequential model.
For linear functions $f$, $g$, and $h$, well-known algorithms such as Kalman filtering and smoothing can be understood as special cases of the sum-product message passing algorithm on this model~\cite{Loeliger2007}, effectively performing Gaussian message passing.
Inspired by the conciseness of~\cite{Loeliger2007}, the aim of the present contribution is to provide tabulated rules for approximate Gaussian message passing through the deterministic nonlinear transformation node depicted in \cref{fig:nonlin-node}, representing the factor $\func{\delta}{y-\fx}$.

To this end, the approximation methods presented in the previous section may now be formulated as message passing on factor graphs, the rules for which are summarized in \cref{tab:mp}.
To obtain these rules, we identify the forward message $\fwd{\mu}_X$ with the filtering distribution $\prob{x|u}$ and the marginal message $\mu_X$ used in the backward pass with the smoothing distribution $\prob{x|u,z}$, with the parameters of the Gaussian distributions defined as in eqs.~\eqref{eq:def-filt}-\eqref{eq:def-smooth}, and $u$ and $z$ denoting available knowledge on $x$ and $y$, respectively, as before.
Eqs.~\eqref{eq:transformed-expectation}-\eqref{eq:numerical-integration} then directly yield the update rules~\eqref{eq:fwd-mean} and~\eqref{eq:fwd-var} for the forward pass.
Eq.~\eqref{eq:rts-results} yields the update rules~\eqref{eq:rts-mean},~\eqref{eq:rts-cov} for the backward pass in the $(\mean_X, \varmat_X)$ parameterization of the Gaussian messages.
This parameterization is, however, not necessarily desirable to use for practical applications since passing backwards through a filtering graph in this parameterization may require multiple matrix inversions for each time slice~\cite{Wadehn2016}.
Therefore, we additionally derive message passing rules in the
\begin{equation}
	\tilde{\infomat}_X = (\fwd{\varmat}_X + \bwd{\varmat}_X)^{-1}, \quad \tilde{\xi}_X = \tilde{\infomat}_X (\fwd{\mean}_X - \bwd{\mean}_X),
\end{equation}
parameterization, which has proven advantageous for efficient realization of the (backwards) smoothing pass on factor graphs such as the one shown in \cref{fig:kalman-fg}~\cite{Wadehn2016}.
Using the Woodbury formula~\cite{Higham2002} and $\varmat_X = (\fwd{\infomat}_X+\bwd{\infomat}_X)^{-1}$~\cite{Loeliger2007}, we obtain
\begin{align*}
	\tilde{\infomat}_X	 &= \fwd{\infomat}_X - \fwd{\infomat}_X \varmat_X \fwd{\infomat}_X \\
	&= \fwd{\infomat}_X - \fwd{\infomat}_X 
	\left( 
		\fwd{\varmat}_X + \fwd{D} (\varmat_Y - \fwd{\varmat}_Y) \fwd{D}^T 
	\right) \fwd{\infomat}_X \\
&= - \fwd{\infomat}_X \fwd{\covmat} \fwd{\infomat}_Y 
	\left(
		- \fwd{\varmat}_Y \tilde{\infomat}_Y \fwd{\varmat}_Y 
	\right) 
	\fwd{\infomat}_Y \fwd{\covmat}^T \fwd{\infomat}_X	 \\
	&= \fwd{\infomat}_X \fwd{\covmat} \tilde{\infomat}_Y \fwd{\covmat}^T \fwd{\infomat}_X,
\end{align*}
corresponding to update rule~\eqref{eq:rts-wtilde}.
For $\tilde{\xi}_X$, using~\cite{Loeliger2007}
\begin{equation*}
	\mean_X = \fwd{\mean}_X - \fwd{\varmat}_X\tilde{\xi}_X = \bwd{\mean}_X + \bwd{\varmat}_X\tilde{\xi}_X,
\end{equation*}
we obtain
\begin{align*}
	\tilde{\xi}_X &= \tilde{\infomat}_X 
	\left( 
		\fwd{\covmat} \tilde{\xi}_Y + \bwd{\varmat}_X \tilde{\xi}_X 
	\right) \\
	&= \tilde{\infomat}_X 
	\left( 
		\fwd{\covmat} \tilde{\xi}_Y + 
		\left(
			\tilde{\infomat}_X^{-1} - \fwd{\varmat}_X 
		\right) 
		\tilde{\xi}_X
	\right) \\
	&= \tilde{\infomat}_X \fwd{\covmat} \tilde{\xi}_Y + \tilde{\xi}_X - \tilde{\infomat}_X \fwd{\varmat}_X \tilde{\xi}_X = \fwd{\infomat}_X \fwd{\covmat} \tilde{\xi}_Y
\end{align*}
as summarized in update rule~\eqref{eq:rts-xitilde}.

\begin{table}[htb]
\fboxsep3mm
\noindent\fbox{%
	\parbox{\linewidth-6mm}{%
	\begin{center}
	\begin{tikzpicture}[FactorNode/.append style = {inner sep = 11pt}]
		\def\xInc {1.7}
		\def\yInc {0.8}
		\SetGraphUnit{1}
		
		\node[FactorNode]    (f)     at ($(0,0)   + (\xInc, 0.0)$)   {${f}$};

		\draw[Edge] (0,0) -- (f)  node[Label, above] {$X$};
		\draw[Edge] (f)  -- ($(f)   + (\xInc, 0.0)$) node[Label, above] {${Y}$};
	\end{tikzpicture}
	\end{center}
	
	\emph{Forward (filtering) pass:} (Using a numerical quadrature procedure with quadrature points $\sigmapoint_i$ and weights $\weight_i$)
	\begin{align}
		\fwd{\mean}_Y & \approx \sum_{i=1}^{\npoints} \weight_i \sfunc{f}{\sigmapoint_i} \label{eq:fwd-mean}\\
		\fwd{\varmat}_Y &\approx \sum_{i=1}^{\npoints} \weight_i (\sfunc{f}{\sigmapoint_i} - \fwd{\mean}_Y)
		(\sfunc{f}{\sigmapoint_i} - \fwd{\mean}_Y)^T
		\label{eq:fwd-var}
	\end{align}
	\emph{Backward (smoothing) pass:}	
	(Valid under the Markov assumption that there is no path from $X$ to $Y$ in the factor graph other than through $f$)
	\begin{align}
		\mean_X & = \fwd{\mean}_X + \fwd{D} (\mean_Y - \fwd{\mean}_Y) \label{eq:rts-mean} \\
		\varmat_X &= \fwd{\varmat}_X + \fwd{D} (\varmat_Y - \fwd{\varmat}_Y) \fwd{D}^T \label{eq:rts-cov}\\
		\tilde{\infomat}_X &= \fwd{\infomat}_X \fwd{\covmat} \tilde{\infomat}_Y \fwd{\covmat}^T\fwd{\infomat}_X \label{eq:rts-wtilde} \\
		\tilde{\xi}_X &= \fwd{\infomat}_X \fwd{\covmat} \tilde{\xi}_Y \label{eq:rts-xitilde}
	\end{align}
	with
	\begin{align}
		\fwd{D} &= \fwd{\covmat} \fwd{\infomat}_Y, = \fwd{\covmat} \fwd{\varmat}_Y^{-1} \label{eq:D-def}\\
		\fwd{\covmat} &\approx \sum_{i=1}^{\npoints} \weight_i (\sigmapoint_i - \fwd{\mean}_X)
		(\sfunc{f}{\sigmapoint_i} - \fwd{\mean}_Y)^T \label{eq:covmat-approx}
	\end{align}
	}
}
	\caption{Approximate Gaussian message passing rules for deterministic nonlinear transformation nodes.}
	\label{tab:mp}
\end{table}

\section{Nonlinear filtering and smoothing}
\label{sec:nonlin-filtering}
The approach described in the previous sections can now be used to describe various new and existing nonlinear filtering and smoothing algorithms by performing forward and backward message passing along factor graphs such as the one shown in \cref{fig:kalman-fg}.
One instance of the class of algorithms that can be derived from this framework is given by the following nonlinear Modified Bryson-Frazier (MBF) smoother for state space models of the form~\eqref{eq:nonlinear-ss} with linear output $h(x) = H x$:
\\\\
\noindent\textbf{Nonlinear MBF Smoother:}
\begin{enumerate}
\item Perform forward message passing using equations~\eqref{eq:fwd-mean} and~\eqref{eq:fwd-var}, as well as the previously proposed update rules (II.1), (II.2), (V.1), and (V.2) from~\cite{Loeliger2016}.
\item Perform backward message passing using equations \eqref{eq:rts-wtilde} and~\eqref{eq:rts-xitilde} as well as the previously proposed update rules (II.6), (II.7), and either (V.4), (V.6), (V.8) or (V.5), (V.7), (V.9) from~\cite{Loeliger2016}.
\end{enumerate}

The smoother is adapted from the factor graph formulation of the MBF smoother~\cite{Bierman1977} for linear systems provided by Loeliger, Bruderer, Malmber, \emph{et al.}~\cite{Loeliger2016}.
It requires just a single matrix inversion for each backward time step and is included here mainly to demonstrate the utility of the presented factor graph representation of nonlinear Gaussian message passing for deriving various nonlinear filters and smoothers.
The smoother may be implemented using any kind of numerical quadrature procedure and amounts to standard message passing on a statistically linearized factor graph~\cite{VanderMerwe2004}.

\section{Conclusion}
In this contribution, local message passing rules for factor graph nodes representing deterministic nonlinear transformations have been derived.
For the forward pass, a linearization is performed using any numerical quadrature method,
and for the backward pass, general Rauch-Tung-Striebel (RTS)-type update rules have been derived in two different message parameterizations.
The resulting message passing rules can be employed in any factor graph
and in particular can be used to perform filtering and smoothing on state space models with state transition, input, and measurement nonlinearities.
Demonstrating the usefulness of transferring results from classical nonlinear filtering theory to the factor graph framework, the modified Bryson-Frazier (MBF) smoother is easily augmented to incorporate nonlinear state transitions and input nonlinearities, requiring only a single matrix inversion in each time step.
In this way, the present contribution adds the capability of handling nonlinear systems to a range of existing algorithms, hence enabling a factor graph description of a whole range of various new and existing algorithms.

\subsection*{Acknowledgements}
The authors would like to thank Maximilian Pilz for interesting discussions on the subject of this article.

\bibliography{references}
\end{document}